\documentclass[sigconf,final]{acmart}
\usepackage{subfigure}
\usepackage{mathrsfs}
\usepackage[ruled,vlined]{algorithm2e}
\usepackage[utf8]{inputenc}
\usepackage{array}
\usepackage{makecell}
\usepackage{booktabs}
\usepackage{multirow}
\usepackage{graphicx}
\usepackage{tabularray}
\usepackage{enumitem}
\setcopyright{acmcopyright}
\usepackage{hyperref}       %
\usepackage{url}            %
\usepackage{nicefrac}       %
\usepackage{microtype}      %
\usepackage[normalem]{ulem}
\usepackage{subcaption}
\usepackage{listings}

\usepackage{amsmath,pifont}
\usepackage[misc]{ifsym}
\usepackage{diagbox}
\usepackage{bigstrut}
\usepackage{adjustbox} %
\usepackage{wrapfig}
\usepackage{caption}
\usepackage{bm}
\usepackage{color}
\usepackage{hyphenat}
\usepackage{balance}
\usepackage[most]{tcolorbox}
\usepackage{mathtools}
\usepackage{xspace}
\usepackage{bbding}

\renewcommand{\thefootnote}{\fnsymbol{footnote}}
\newcommand{\fix}{\color{black}}

\AtBeginDocument{%
  }

\setcopyright{acmlicensed}
\copyrightyear{2018}
\acmYear{2018}
\acmDOI{XXXXXXX.XXXXXXX}

\acmConference[Conference acronym 'XX]{Make sure to enter the correct
  conference title from your rights confirmation emai}{June 03--05,
  2018}{Woodstock, NY}
\acmISBN{978-1-4503-XXXX-X/18/06}

\begin{document}

\title{CityGPT: Empowering Urban Spatial Cognition of Large Language Models}

\author{Jie Feng*}
\affiliation{%
 \institution{Department of Electronic Engineering, BNRist,\\Tsinghua University}
 \country{Beijing, China}
}
\email{fengj12ee@hotmail.com}

\author{Tianhui Liu*}
\affiliation{%
 \institution{School of Electronic and Information Engineering,\\Beijing Jiaotong University}
 \country{Beijing, China}
}
\email{21211125@bjtu.edu.cn}

\author{Yuwei Du}
\affiliation{%
 \institution{Department of Electronic Engineering, BNRist,\\Tsinghua University}
 \country{Beijing, China}
}
\email{duyw23@mails.tsinghua.edu.cn}

\author{Siqi Guo}
\affiliation{%
 \institution{Department of Electronic Engineering,\\Tsinghua University}
 \country{Beijing, China}
}
\email{guosq21@mails.tsinghua.edu.cn}

\author{Yuming Lin}
\affiliation{%
 \institution{Department of Urban Planning,\\Tsinghua University}
 \country{Beijing, China}
}
\email{linyuming9@mail.tsinghua.edu.cn}

\author{Yong Li$\dagger$}
\affiliation{%
 \institution{Department of Electronic Engineering, BNRist,\\Tsinghua University}
 \country{Beijing, China}
}
\email{liyong07@tsinghua.edu.cn}

\renewcommand{\shortauthors}{Jie Feng et al.}

\begin{abstract}
Large language models(LLMs), with their powerful language generation and reasoning capabilities, have already achieved notable success in many domains, e.g., math and code generation. However, they often fall short when tackling real-life geospatial tasks within urban environments. This limitation stems from a lack of physical world knowledge and relevant data during training. To address this gap, we propose \textit{CityGPT}, a systematic framework designed to enhance LLMs' understanding of urban space and improve their ability to solve the related urban tasks by integrating a city-scale `world model' into the model. Firstly, we construct a diverse instruction tuning dataset, \textit{CityInstruction}, for injecting urban knowledge into LLMs and effectively boosting their spatial reasoning capabilities. Using a combination of \textit{CityInstruction} and open source general instruction data, we introduce a novel and easy-to-use self-weighted fine-tuning method (\textit{SWFT}) to train various LLMs (including ChatGLM3-6B, Llama3-8B, and Qwen2.5-7B) to enhance their urban spatial capabilities without compromising, or even improving, their general abilities. Finally, to validate the effectiveness of our proposed framework, we develop a comprehensive text-based spatial benchmark \textit{CityEval} for evaluating the performance of LLMs across a wide range of urban scenarios and geospatial tasks. Extensive evaluation results demonstrate that smaller LLMs trained with \textit{CityInstruction} by \textit{SWFT} method can achieve performance that is competitive with, and in some cases superior to, proprietary LLMs when assessed using \textit{CityEval}. Our work highlights the potential for integrating spatial knowledge into LLMs, thereby expanding their spatial cognition abilities and applicability to the real-world physical environments. The dataset, benchmark, and source code are open-sourced and can be accessed through \url{https://github.com/tsinghua-fib-lab/CityGPT}.
\end{abstract}

\maketitle

\footnotetext[1]{These authors contributed equally.}
\footnotetext[2]{Corresponding author, email: liyong07@tsinghua.edu.cn}

\renewcommand{\thefootnote}{\arabic{footnote}}
\setcounter{footnote}{0}

\section{Introduction}\label{sec:intro}

In recent years, large language models(LLMs) have made rapid advancements across various language-based scenarios, such as chat~\cite{brown2020language} and code generation~\cite{achiam2023gpt}. Multiple studies~\cite{wei2022emergent} have shown that LLMs exhibit powerful generalization across a wide range of tasks and demonstrate impressive reasoning ability over complex tasks. These developments have significantly contributed to the progress of general artificial intelligence and have encouraged the broader application of LLMs across diverse domains. Consequently, various domain-specific LLMs like BloombergGPT~\cite{wu2023bloomberggpt} for finance, Med-PaLM~\cite{singhal2023large} for medicine and Llemma~\cite{azerbayev2023llemma} for math have been proposed and have achieved promising results.

\begin{figure*}[!htbp]
    \centering
    \includegraphics[width=1\textwidth]{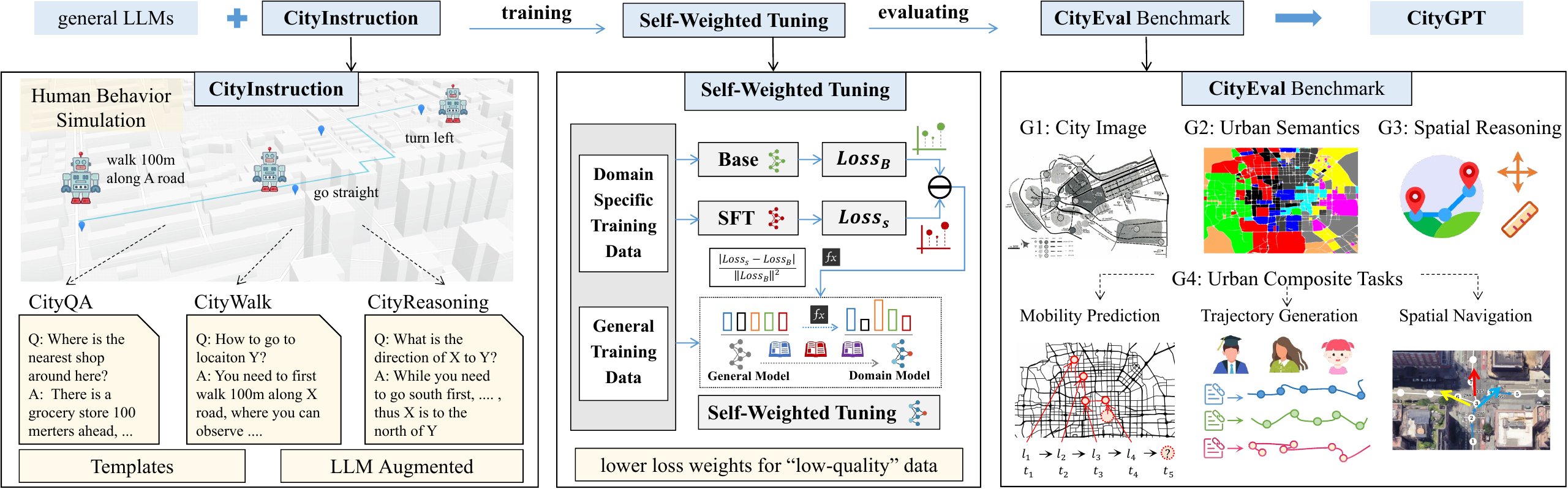}
    \caption{An overview of \textit{CityGPT}, including \textit{CityInstruction} dataset, self-weighted tuning \textit{SWFT} method, and \textit{CityEval} benchmark. \textit{CityInstruction} comprises \textit{CityQA}, \textit{CityWalk} and \textit{CityReasoning}, while \textit{CityEval} includes \textit{City Image}, \textit{Urban Semantics}, \textit{Spatial Reasoning} and \textit{Composite Tasks}.}
    \label{fig:framework}
\end{figure*}

In geography and urban science field, researchers also investigate the potential of LLMs in domain-specific tasks like geospatial understanding and prediction tasks~\cite{gurnee2023language, manvi2023geollm, roberts2023gpt4geo, manvi2024large, bhandari2023large, godey2024scaling, mai2023opportunities}. For example, Gurnee et al.~\cite{gurnee2023language} investigate weather Llama2~\cite{touvron2023llama} really learns the map of the world by training linear regression probes to predict the real location and Manvi et al.~\cite{manvi2023geollm} design prompts to extract geospatial knowledge from LLM for downstream prediction tasks. In the global scale or national scale, researchers find that LLMs are good at representing the coarse location and related geospatial knowledge like demographic indicators. However, they also find that LLMs become struggling when the geospatial task breaks down to the city scale, e.g., the location coordinates prediction accuracy dropped from more than 70\% for cities in USA to less than 30\% for Point of Interests (PoIs) in New York~\cite{gurnee2023language}. The results indicate that after training on the online web text, LLMs may lack the detailed geospatial knowledge of the offline physical world in the city. Besides, existing evaluations have deficiencies in two aspects which greatly limits understanding the utility of LLMs in urban space. On the one hand, most of the evaluation tasks are based on the simple location coordinates which is just a small part of the space, more concepts and tasks need to be validate, e.g., fundamental elements in the image of the city~\cite{lynch1964image}. On the other hand, most of the evaluation are conducted in the global scale or national scale, limited results in the city scale are available. The question of whether LLMs can truly be applied to solve the geospatial task in the city scale and whether they own similar spatial cognition like human remains unknown.

In this paper, we propose \textit{CityGPT}, a systematic framework for evaluating and enhancing the capability of LLMs on understanding the urban space and solving the urban geospatial tasks. As the first component of \textit{CityGPT}, we construct a instruction tuning dataset \textit{CityInstruction}, which is diverse and effective for enhancing the capability of general LLMs on understanding the urban space. We follow the similar experience of human exploring and perceiving the urban space in the daily life via a mobility simulator with real map to build the dataset. Furthermore, we extend the experience dataset by generating explicit intermediary spatial reasoning steps for high level urban tasks which encourage the model to learn the general reasoning paradigms in urban space. 
As the second component of \textit{CityGPT}, we propose a robust self-weighted fine-tuning method \textit{SWFT} that automatically assesses the quality of domain data, reweights the loss accordingly, and effectively enhances the spatial skills of LLMs, while minimizing negative impact on their general performance. 
We first train a warm-up LLM using standard SFT with the entire CityInstruction dataset. Then, we assess the quality of each example based on the evaluation losses of both the base LLM and the warm-up LLM. Based on observations of high-quality data and loss variations, we propose self-weighted fine-tuning method, which assigns smaller learning weights in the loss function to 'low-quality' data, ensuring more robust and effective knowledge learning.
As the third component of \textit{CityGPT}, we build a comprehensive evaluation benchmark, \textit{CityEval}, to evaluate the capability of LLMs on various urban scenarios and downstream tasks. Follow the experience from urban planning~\cite{lynch1964image}, neurocognitive science~\cite{epstein2017cognitive} and geoscience~\cite{mai2023opportunities}, the evaluation task in \textit{CityEval} is divided into four groups: \textit{City Image} task group for measuring the intuitive understanding of urban fundamental elements, \textit{Urban Semantics} task group for understanding the effects of human activities on urban environment, \textit{Spatial Reasoning} task group for high level spatial cognitive capability evaluation, \textit{Composite Task} group for evaluating the integrated capability of LLMs, which includes mobility prediction~\cite{wang2023would}, behavior generation~\cite{shao2024beyond} and street navigation~\cite{coutrot2022entropy} with more complicated context and instructions. In summary, our contributions are as follows,

\begin{itemize}[leftmargin=1.5em,itemsep=0pt,parsep=0.2em,topsep=0.0em,partopsep=0.0em]
\item To our best knowledge, \textit{CityGPT} is the first systematic framework to evaluate and enhance the spatial cognition abilities of general LLMs in the urban environment.
\item We propose a mobility behavior simulation based instruction tuning data synthesis method which could prepare high-quality data \textit{CityInstruction} for injecting urban knowledge into LLMs.
\item We propose an effective fine-tuning method, self-weighted fine-tuning (\textit{SWFT}), to enable robust and easy-to-use domain-specific model training without compromising general capabilities.
\item We propose \textit{CityEval}, a comprehensive urban spatial evaluation benchmark to accessing the performance of LLMs on urban spatial knowledge and reasoning abilities. 
\item Extensive experiments on \textit{CityEval} show that our method effectively enhance the spatial knowledge and capabilities of general LLMs. After training, smaller models achieve performance comparable to or even better than top proprietary LLMs.
\end{itemize}

\section{Methods}

In this paper, we propose a systematic framework to evaluate and enhance the capability of LLMs on urban tasks and applications. The whole framework is shown in Fig.~\ref{fig:framework}, which comprises three central components: \textit{CityInstruction}, \textit{Self-weighted Fine-Tuning}, and \textit{CityEval}. 
We first introduce \textit{CityInstruction} in Section~\ref{sec:CityInstruction}, a dataset designed to inject urban knowledge into general LLMs, enhancing their ability to handle urban-related tasks.
Next, in Section~\ref{sec:tuning}, we introduce \textit{self-weighted fine-tuning}, an effective and user-friendly instruction-tuning method that enables \textit{CityGPT} to be trained efficiently and robustly across diverse LLMs using mixed domain specific data.
Finally, in Section~\ref{sec:cityeval}, we introduce \textit{CityEval}, a benchmark designed to comprehensively evaluate LLMs on their understanding of urban spaces and ability to solve urban tasks.

\subsection{\textit{CityInstruction} Construction} \label{sec:CityInstruction}

As mentioned before, general LLMs struggle to solve the task in the city scale due the lack of offline urban knowledge which is rare in the online web text. A naive approach to compensate for the shortcoming is to learn the urban knowledge directly, e.g., raw map data. However the raw map data which is designed for efficient storage and computing is not friendly for learning. Some studies have been done to solve this problem and efficiently ground the geospatial information and natural language. For example, Huang et al.~\cite{huang2022ernie} propose to utilize online user logs to construct heterogeneous graph and sample random walk to construct the sequence data to train a BERT model. However, the online user log is not public available which make the former method disabled for the public. Besides, GeoLM~\cite{li2023geolm} use geospatial entity from the Wikipedia to align with the same entity in open street map and validate the effectiveness in several downstream knowledge graph task like toponym linking and relation extraction. However, the entity recorded in the online Wikipedia is very limited which restricts the application of this method in the real world.

To solve this problem, we revisit the mechanisms of spatial cognition in humans. Actually, human beings perceive and recognize the physical world through embodied experiences, often in the form of multi-view data, during daily activities such as commuting and wandering. Thus, we propose to simulate the human mobility behaviors in the daily life to collect multi-view data simultaneously as the embodied experience data for model tuning. For example, the embodied experience data can be, "What can I observe at the current location?", the answer may include the information of PoI, road, AoI and so on. Based on the collected simple experience data, we first manually design templates for each kind of data and then utilize ChatGPT~\cite{achiam2023gpt} with carefully designed prompts to extend these templates with more diverse formats. In this way, we obtain the diverse multi-view experience data in the forms of instructions, including the \textit{CityQA} and \textit{CityWalk}. Finally, we construct the \textit{CityReasoning} dataset with intermediary steps by solving spatial reasoning problems with manually designed solutions and ChatGPT assisted solutions. For example, the spatial reasoning problem can be inferring the spatial orientation and distance relationships between locations based on the navigation routing as input. 

\subsubsection{CityQA-Single Step Exploration}
In \textit{CityQA}, we define simple questions for all kinds of single entity in the urban space. The entities considered in the dataset are PoI, AoI, road and junction of roads. For example, we construct question about the basic information of PoI, including its category, address and coordinates, and also relations with other nearby entities. The question of AoI is similar to the PoI. For the road and junction, only relation base questions are considered. As mentioned before, each kind of question have at least 3 templates which are manually defined and rewriting with the help of ChatGPT for diversity. It is noted that due to a large portion of addresses of entities being absent, we reconstruct the address of them based on the road network. In this way, we can directly link the major entities in the same urban space with the same address representation via road network. To some extent, this alignment method via road network based address is similar to the entity alignment methods from~\cite{li2023geolm} when LLMs are highly skilled at recognizing and aligning similar entity~\cite{xu2023large}. Meanwhile, proposed alignment method are easy to implement and can be widely-used for any regions around the world.

\begin{figure}
    \centering
    \includegraphics[width=0.6\linewidth]{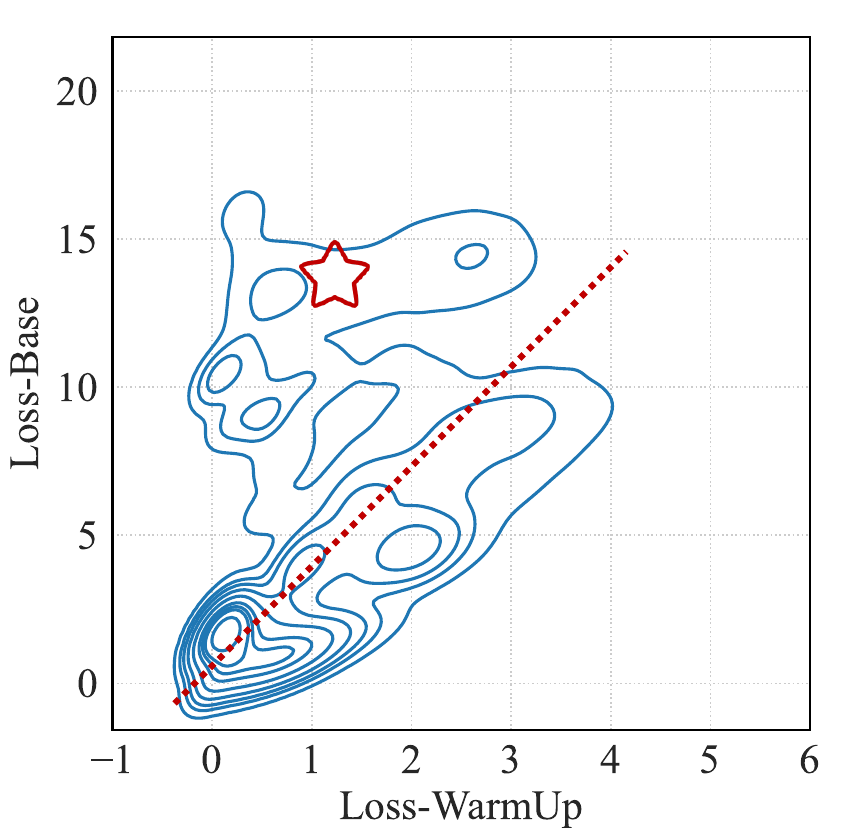}
    \caption{The loss of data samples before and after training, where the dashed trend line represents the average learning ratio, and red pentagrams highlight the anomalous region.}
    \vspace{-0.3cm}
    \label{fig:loss-insights}
\end{figure}

\subsubsection{CityWalk-Multi Step Exploration}
\textit{CityQA} covers the single entity and its nearby relations, which can be regarded as a single step \textit{CityWalk} with random explored positions. \textit{CityWalk} with a long-term temporal and spatial window can provide us more similar and diverse embodied experiences of urban space than human beings. Here, following the practices from embodied agents for house-holding~\cite{li2022pre,xiang2024language}, we design two work modes to drive a data collection agent to construct the \textit{CityWalk} dataset by exploring the urban space in the simulator. In the first mode, given a starting point and a goal, the agent freely explores the urban space until the goal is reached. For examples, when the agent is required to start from its predefined home to buy some vegetables back, it can collect potential experience data by interacting with the simulator with APIs like search, walk, and so on. While this mode is intuitive, it requires much more carefully designed post-processing efforts. In second mode, we directly assign task with predefined origin and destination to the agent which only needs to follow the fixed path to collect multi-step data including the routing between locations and other observations during the trip. We choose the second mode as the default data collection mechanism when this mode works controllably and efficiently. The first mode is only used to generate small portion data for diversity.

\subsubsection{CityReasoning-Exploration with Explicit Spatial Reasoning Steps} \label{method:cityreasoning}
Spatial reasoning problem in the urban space are not widely studied in the existing works when most of them mainly focus on the reasoning based on geographic coordinates~\cite{bhandari2023large,godey2024scaling}. However, geographic coordinates based reasoning is not consistent with the mechanism of spatial cognition in humans, who usually engage in spatial reasoning by integrating the approximate memories of visited path with rough calculations. Thus, we construct \textit{CityReasoning} dataset whose reasoning mechanism is aligned with human cognitive habits to effectively enhance the spatial reasoning capability of LLMs. Here, we focus on most basic direction and distance based spatial reasoning problems. Most topology related problems in our setting can be transferred to these two questions. To construct \textit{CityReasoning}, we first random select two locations and obtain their navigation path as context. Based on the navigation path, we can generate the explicit reasoning steps of inferring direction and distance and translate them with predefined templates into the format of instruction data. For the distance reasoning problem, the intermediate reasoning process is to perform a rough summation of the traversed path step by step. For the direction reasoning problem, the intermediate reasoning process is to group the movement by the direction and then compare their values to obtain the results.

\subsection{\textit{SWFT}: Self-Weighted Fine-Tuning} \label{sec:tuning}
\subsubsection{General Instruction Data}
Following the practice of previous works~\cite{zeng2023agenttuning, dong2023abilities}, we mix the \textit{CityInstruction} with the diverse general instruction data during the instruction tuning stage for reducing the risk of catastrophic forgetting of general capability. We introduce four general datasets in our experiments, ShareGPT~\footnote{\url{https://huggingface.co/datasets/shareAI/ShareGPT-Chinese-English-90k}}, UltralChat~\cite{ding2023enhancing}, Open-Platypus~\cite{lee2023platypus} and AgentTuning~\cite{zeng2023agenttuning}. Besides, we introduce geographical datasets GeoGLUE~\cite{li2023geoglue} to enhance the geographical language understanding capability and language based spatial reasoning dataset StepGame~\cite{shi2022stepgame} and ReSQ~\cite{mirzaee2022transfer} for enhancing the qualitative spatial reasoning capability of \textit{CityGPT} via language. While GeoGLUE does not follow the instruction format, we design simple templates to convert the data. We also utilize GPT-4 as the automatic data annotator and data quality evaluator to generate intermediate reasoning steps of StepGame and ReSQ to reduce the learning difficulty of small LLMs.

\subsubsection{Self-Weighted Fine-Tuning}

Even when mixed with general data, the trained model still performs poorly to some extent, exhibiting a decline in general capabilities and a degradation in domain-specific performance. To investigate this issue, as shown in Figure~\ref{fig:loss-insights}, we analyzed the loss of \textit{CityInstruction} data samples before and after training and identified a subset of anomalous data points, highlighted as red pentagrams in the figure. These data points have an exceptionally high loss on the base model (e.g., Llama3-8B, $M_{base}$), indicating that they are either of poor quality or introduce entirely new knowledge that significantly deviates from the base model’s prior distribution. Furthermore, after training, the loss reduction for these data points remains smaller than the average reduction observed across the entire dataset (as indicated by the dashed trend line). This confirms that these data points are indeed of 'low quality' and difficult for the model to learn from, potentially harming overall performance. Thus, based on these observations, we propose the \textit{self-weighted fine-tuning (SWFT)} method, which leverages learning dynamics to generate personalized weights for each data sample using a function $f$, thereby adaptively adjusting sample-specific weights $w$ in the loss function. Specifically, the loss function $\mathcal{L}_{swft}$ of proposed \textit{SWFT} for model $M$ is formulated as follows,
\begin{align*}
    \quad & \mathcal{L}_{\text{swft}}(M) = - \frac{1}{N} \sum_{i=1}^{N} w_i \sum_{t=1}^{T}  \log P_M(y_t | x, y_{<t}), \\
    \quad & w_i = f(\mathcal{L}_i(M_{\text{base}}), \mathcal{L}_i(M_{\text{warm}})), \\
    \quad & f_{exp} = \frac{|\mathcal{L}(M_{\text{warm}}) - \mathcal{L}(M_{\text{base}})|}{\|\mathcal{L}(M_{\text{base}})\|_2},
\end{align*}
\noindent
where $f_{exp}$ denotes an representative function $f$ by considering the trend in the observation, $x$ is the input token, $T$ is the token length of data instance $i$, $y_t$ is the next token, $N$ is the number of data samples, $\mathcal{L}$ denotes the normal cross entropy loss.  
According to the above formula, "low-quality" data—characterized by a large initial loss and minimal loss reduction—will automatically be assigned lower weights during tuning, as determined by the base LLM $M_{base}$ and the warm-up trained LLMs $M_{warm}$. This allows the model to effectively focus more on well-matched, high-quality data, leading to improved and more robust performance. We believe $f_{exp}$ is not the only possible weighting function. Here, we define $f_{exp}$ in the simplest form due to the limitation of computation.

\subsection{\textit{CityEval} Benchmark} \label{sec:cityeval}
\begin{figure}
   \centering
    \includegraphics[width=0.45\textwidth]{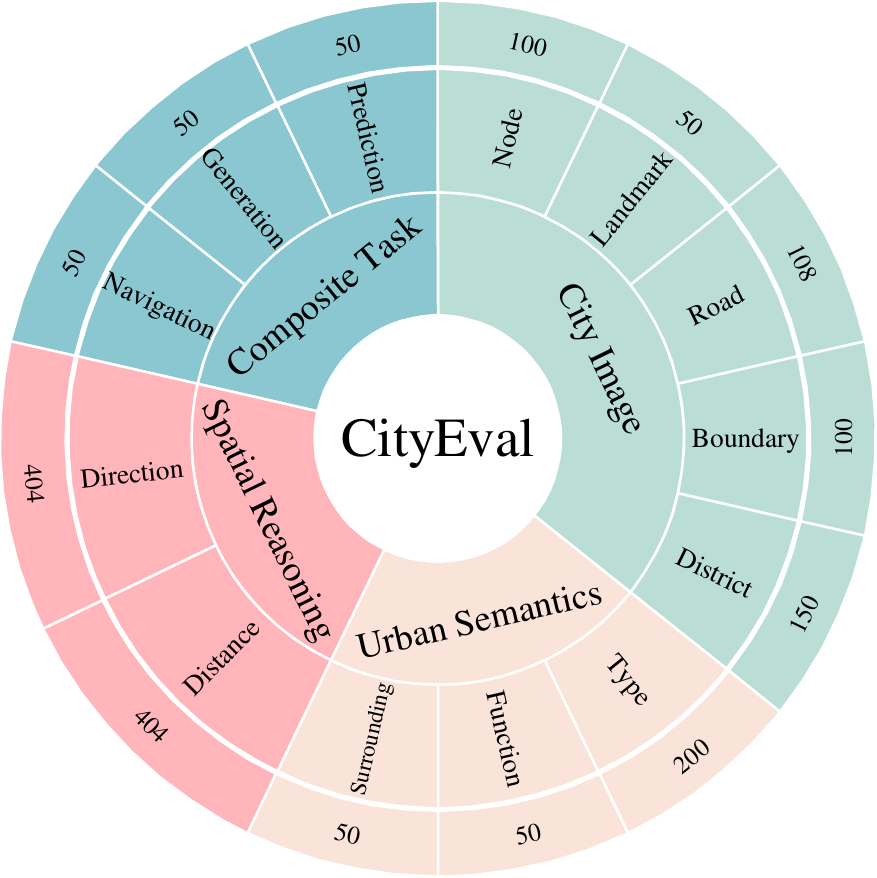}
    \caption{\fix Composition of \textit{CityEval}.}
    \label{fig:eval-class}
\end{figure}

Different form existing works~\cite{mai2023opportunities, roberts2023gpt4geo, gurnee2023language} which mainly focus on evaluation on the global scale or national scale, we propose a systematic evaluation benchmark \textit{CityEval} to testify the capability of LLMs in urban space. Following the common experiences from different fields~\cite{lynch1964image, epstein2017cognitive, mai2023opportunities}, as Fig.~\ref{fig:eval-class} shows, our benchmark contains four sub-modules with emphasizing different aspects of spatial cognition of human and applications of urban science, including \textit{City Image}, \textit{Urban Semantics}, \textit{Spatial Reasoning} and \textit{Composite Tasks}. 

\subsubsection{City Image-Fundamental Elements of The City In the Mind}
Lynch~\cite{lynch1964image} explored the mechanism of human perceiving and remembering the urban environments and found that human usually deconstruct urban space into a combination of five fundamental elements: paths, edges, districts, nodes, and landmarks in their mind. Similar concepts and mechanism are also verified by the neurocognitive science~\cite{epstein2017cognitive}. Based on these observations, we propose the \textit{City Image} task group to evaluate whether LLMs can understand these concepts and solve the related questions. In \textit{City Image}, we manually design various questions for each element, e.g., attributes and underlying relations. For example, while people use path in the mind to remember how to arrive different locations, one typical question for it is "Do you know the origin and destination junction of Nanyuan road?" It is noted that this question dose not exist in the instruction tuning data directly, model needs to learn from the experiences in \textit{CityWalk} and answer this question by organizing its memory about all the possible junctions of Nanyuan road. Another example is about the concept of boundary. While roads usually play the role of separating different areas in the city, we design the question like "Which of the following roads serve as the boundaries of region X?" to confirm weather the model understands the concept of boundary in the urban space. In summary, we design 12 types of questions for 5 fundamental elements to construct the \textit{City Image}. 

\subsubsection{Urban Semantics-Human Activities and Urban Environment}
Different from the \textit{City Image}, \textit{Urban Semantics} pay attention to the human activities happened in the urban environment and evaluate the capacity of understanding the urban functions. The understanding and prediction of urban functions is the fundamental task of geoscience~\cite{mai2023opportunities} and urban science~\cite{yuan2012discovering}. In simple terms, we define the task of \textit{Urban Semantics} as inferring the functions of areas with knowing the PoI distribution and the most likely missing PoIs in the environment. For example, the question about inferring the functions of areas can be "There are PoI A, PoI B, ..., PoI G in the area Y, what is the potential function of area Y?". In summary, we design 6 types of questions to construct the \textit{Urban Semantics}. 

\subsubsection{Spatial Reasoning-Reasoning in Urban Space}
Compared with the former two tasks which require more about the capacity of memorization and association analysis, we introduce \textit{Space Reasoning} task group to evaluate the capacity of quantitative reasoning and spatial cognition which is more challenging for LLMs. The design of \textit{Spatial Reasoning} is similar to the methods in Section~\ref{method:cityreasoning} \textit{CityReasoning}. Here, we set two testing scenarios for the spatial reasoning evaluation. The first testing scenario is whether the question contains the necessary context for spatial reasoning. When no necessary context is provided, model needs to reasoning over its memory which is a more challenging process. The second testing scenario is whether the urban space of reasoning has been saw by the model in the training. Besides, when evaluating models trained by \textit{CityReasoning} introduced before, we carefully check the evaluation data to prevent the potential data leakage. In summary, we design 20 types of questions for \textit{Spatial Reasoning}.

\subsubsection{Urban Composite Tasks-Solving Realistic Urban Tasks}
Finally, we introduce the \textit{Composite Tasks} group which consists of mobility prediction~\cite{wang2023would}, trajectory generation~\cite{shao2024beyond} and spatial navigation~\cite{coutrot2022entropy} for assessing the integration capability of LLMs in urban space. All these tasks require diverse capability of LLMs to understand the urban environments and human behavior to complete. Detailed introduction of these tasks can refer to the appendix.

\section{Experiments} \label{Experiments}

\begin{table*}[ht]
\centering
\caption{Main results on \textit{CityEval}. CityGPT-7B significantly outperforms than baselines in most tasks. `CI' denotes city image, `US' denotes urban semantics, and `SR' denotes spatial reasoning in \textit{CityEval} benchmark.}
\label{table:main}
\resizebox{1\textwidth}{!}{
\begin{tabular}{lcccccccccccc} 
\toprule
\textbf{City}& \multicolumn{3}{c}{\textbf{Beijing}} & \multicolumn{3}{c}{\textbf{London}} & \multicolumn{3}{c}{\textbf{NewYork}} &  & \textbf{Paris} &  \\
\textbf{Tasks} & \multicolumn{1}{c}{\textbf{CI}$\uparrow$} & \textbf{US}$\uparrow$ & \textbf{SR}$\uparrow$ & \textbf{CI}$\uparrow$ & \textbf{US}$\uparrow$ & \textbf{SR}$\uparrow$ & \textbf{CI}$\uparrow$ & \textbf{US}$\uparrow$ & \textbf{SR}$\uparrow$ & \textbf{CI}$\uparrow$ & \textbf{US}$\uparrow$ & \textbf{SR}$\uparrow$ \\ 
\cmidrule{2-13}
\textbf{ChatGLM3-6B} & 0.300 & 0.477 & 0.248 & 0.269 & 0.450 & 0.241 & 0.315 & 0.430 & 0.254 & 0.291 & 0.383 & 0.263 \\
\textbf{Qwen2.5-7B}& 0.325 
& 0.587 
& 0.164 
& 0.303 & 0.480 & 0.256 
& 0.311 & 0.503 & 0.256 
& 0.277 & 0.483 & 0.232 
\\
\textbf{LLama3-8B} & 0.286 & 0.520 & 0.285 & 0.312 & 0.490 & 0.282 
& 0.331 & 0.503 & 0.277 
& 0.265 & 0.463 & 0.281 
\\ 
\textbf{Gemma2-9B} & 0.097 
& 0.207 
& 0.159 
& 0.168 & 0.220 & 0.214 
& 0.168 & 0.197 & 0.209 
& 0.162 & 0.197 & 0.191 
\\
\midrule
\textbf{Mistral-Small-24B} & 0.288 
& 0.587 
& 0.293 
& 0.315 & 0.540 & 0.339 
& 0.332 & 0.580 & 0.317 
& 0.342 & 0.493 & 0.320 
\\
\textbf{Gemma2-27B} & 0.326 
& 0.573 
& 0.225 
& 0.363 & 0.520 & 0.245 
& 0.346 & 0.557 & 0.238 
& 0.372 & 0.490 & 0.235 
\\ 
\textbf{Qwen2.5-32B} & 0.400 
& \textbf{0.627}& 0.405 
& 0.360 & 0.540 & 0.419 
& 0.383 & 0.543 & 0.390 
& 0.337 & 0.503 & 0.391 
\\ 

\textbf{LLama3.1-70B} & 0.357 
& 0.570 
& 0.338 
& 0.372 & 0.540 & 0.393 
& 0.412 & \uline{0.597}& 0.377 
& 0.388 & 0.473 & 0.377 
\\ 
\textbf{Qwen2.5-72B} & 0.408 
& 0.607 
& 0.293 
& 0.345 & 0.550 & 0.363 
& \uline{0.417}& 0.577 & 0.319 
& 0.346 & 0.453 & 0.336 
\\ 
\textbf{LLama3.1-405B} & \uline{0.417}& 0.587 
& \uline{0.414}& \uline{0.451}& \uline{0.593}& \uline{0.484}& 0.400 & 0.580 & \uline{0.424}& \textbf{0.457}& \uline{0.513}& \uline{0.467}\\ 

\midrule
\textbf{GPT-3.5} & 0.269 
& 0.553 
& 0.249 
& 0.297 & 0.510 & 0.272 
& 0.354 & 0.490 & 0.239 
& 0.299 & 0.473 & 0.275 
\\
\textbf{GPT-4omini} & 0.319 & 0.533 & 0.246 & 0.386 & 0.513 & 0.308 & 0.369 & 0.523 & 0.291 & 0.369 & 0.503 & 0.293 \\
\midrule
\textbf{CityGPT-Qwen2.5-7B}& \textbf{0.502}& \uline{0.620}& \textbf{0.552}& \textbf{0.525}& \textbf{0.597}& \textbf{0.603}& \textbf{0.485}& \textbf{0.607}& \textbf{0.585}& \uline{0.439}& \textbf{0.527}& \textbf{0.619}\\
\textbf{ vs. Qwen2.5-7B}& +54.50\%& +5.68\%& +236.59\%& +73.08\%& +24.31\%& +135.55\%& +55.92\%& +20.54\%& +128.52\%& +58.36\%& +8.98\%& +166.81\%\\
\textbf{ vs. Best Baseline} & +20.29\%& -1.07\%& +33.33\%& +16.37\%& +0.57\%& +24.59\%& +16.24\%& +1.68\%& +37.97\%& -4.03\%& +2.61\%& +32.55\% \\
\bottomrule
\end{tabular}}
\label{table:main}
\end{table*}

\subsection{Settings}
\subsubsection{General Evaluation Tasks} 
In addition to the proposed \textit{CityEval} benchmark, we also evaluate our model using general benchmarks, including MMLU~\cite{hendrycks2020measuring} to assess general knowledge abilities, GSM8K~\cite{cobbe2021training} to evaluate mathematical abilities, and BBH~\cite{suzgun2022challenging} to evaluate common sense reasoning abilities.

\subsubsection{Evaluation Metrics} 
For general evaluation tasks including MMLU, BBH and GSM8K, we use opencompass~\cite{2023opencompass} with default generation settings as the evaluation tool to calculate the score. For the first three types of tasks in \textit{CityEval}, the questions are organized as single-choice questions with at least 4 choices (up to 10 choices ) and accuracy is chosen as metric. For the last composite application tasks in \textit{CityEval}, we provide all the models with 1-shot example and use their common practices to define metrics.

\subsubsection{Evaluation Cities} 
We choose Beijing, London, NewYork and Paris and perform the experiments across the entire geographical areas of these cities. To conduct the out-of-domain validation, SanFrancisco does not participate in the evaluation but only generates task instruction data for other cities. 

\subsubsection{LLM Baselines} We consider the following baselines and divide them into 3 groups: small open source LLMs group with about 7B parameters including ChatGLM3-6B~\cite{zeng2022glm}, Qwen2.5-7B~\cite{qwen2.5}, Llama3-8B~\cite{llama3-meta} and Gemma2-9B~\cite{gemma_2024}, large open source LLMs group with about 100B parameters including Mistral-Small-24B~\cite{mistral-small-3}, Gemma2-27B~\cite{gemma_2024}, Qwen2.5-32B~\cite{qwen2.5}, LLama3.1-70B~\cite{Dubey2024TheL3}, Qwen2.5-72B~\cite{qwen2.5} and LLama3.1-405B~\cite{Dubey2024TheL3}, and commercial APIs group including GPT-3.5 and GPT-4omini~\cite{achiam2023gpt}. For open source LLMs, we deploy them with the help of vllm~\cite{kwon2023efficient}. The max output token are set to 500, the repetition penalty is set to 1.0, the temperature is set as 0, and it is changed to 1.0 when needing sample answers.

\begin{figure}
\centering
\includegraphics[width=0.48\textwidth]{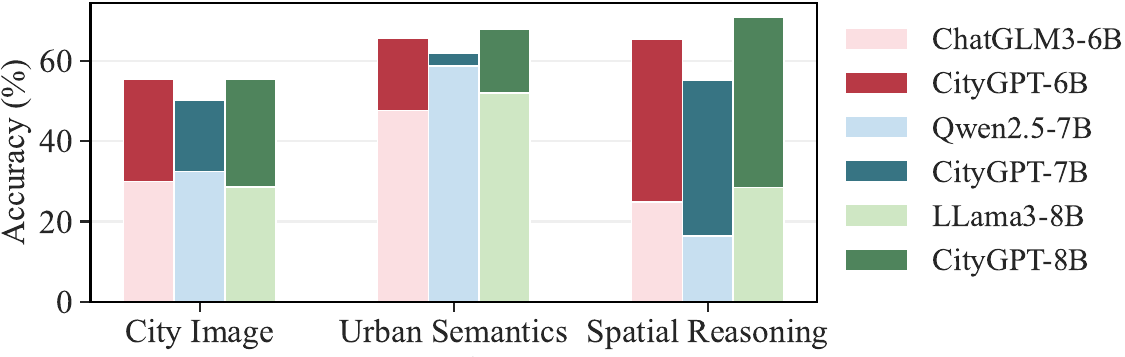}
\caption{\fix The performance of CityGPT@Beijing consistently exceeds that of the baseline across various base models on CityEval benchmark.}
\label{fig:different_base_result}
\end{figure}

\subsection{Overall Performance on CityEval}
The main results of CityGPT on four cities are presented in Table~\ref{table:main}. Here, we use Qwen2.5-7B as the start point and use the proposed \textit{CityInstruction} to obtain the final model CityGPT-7B. It is noted that we only report the performance of top 3 tasks in \textit{CityEval} whose metrics is the accuracy. Results on the final composite task are presented in section~\ref{exp:app}.

We first analyze the results from Beijing. In the first 7B parameter group, we find that more advanced LLMs can achieve better performance in all the tasks, e.g., LLama3-8B outperforms the older ChatGLM3-6B around 23\% in City Image task. It is worth noting that the newer Gemma2-9B unexpectedly shows the weakest performance. We speculate that this is due to the distribution of its training data, which makes it unsuitable for urban-related tasks. Performance of larger parameter group (about 100B) continues to outperforms than small parameter group about 0.69\%-330\%, with the largest open source model LLama3.1-405B achieving the best performance in 2 tasks, even surpassing the GPT series. The results demonstrate the effectiveness of proposed evaluation benchmark \textit{CityEval}. Meanwhile, we notice that there is great potential for overall improvement in the benchmark when the best performance among all the task is only 0.627. Overall, CityGPT-Qwen2.5-7B exhibits significant improvement over all baselines in most of tasks. Compared with the best baseline, the performance gain of CityGPT-7B is 33.33\% in the challenging Spatial Reasoning task. And it outperforms its original version Qwen2.5-7B with at least 5.68\% in the urban semantic task and 236.59\% in the spatial reasoning tasks. These results present the effectiveness of proposed instruction tuning dataset \textit{CityInstruction} which successfully teach the small LLMs with various capabilities in the urban space and achieve outstanding performance than all the powerful general LLMs including the commercial API. 

The results from London, New York, and Paris are similar to those from Beijing. LLama3.1-405B is generally the best-performing baseline and even surpasses CityGPT-7B in the City Image task in Paris. However, apart from this, our CityGPT-7B outperforms all other models across all tasks. 
As shown in the Figure~\ref{fig:different_base_result}, this illustrates the performance of different base models(including ChatGLM3-6B, Qwen2.5-7B and LLama3-8B) on \textit{CityEval} after being trained with \textit{CityInstruction} in Beijing. All CityGPT models outperform the base models across all tasks. This performance gap is particularly evident in the most challenging Spatial Reasoning Task, where CityGPT achieves 148\% to 237\% higher accuracy compared to the base models. 
In summary, proposed evaluation benchmark \textit{CityEval} effectively distinguishes the diverse capabilities of different LLMs in understanding urban space, and \textit{CityInstruction} can be used to effectively improve the performance of smaller LLMs in urban capabilities.

\subsection{General Capabilities and Self-Weighted Fine-Tuning of CityGPT} \label{exp:SWFT}
In this section, we first evaluate the performance of CityGPT in general benchmark. Table~\ref{table:general} presents the performance of CityGPT trained on different base models, which indicates that CityGPT models are generally comparable to the base models in terms of general capabilities. CityGPT-Qwen2.5-7B shows improvements on MMLU and BBH, while maintaining similar performance on GSM8K. However, there are notable decreases on MMLU and GSM8K with CityGPT-LLama3-8B, a decline is also observed on BBH with CityGPT-ChatGLM-6B. Therefore, we propose a Self-Weighted Fine-Tuning approach to address these issues.

\begin{table}[ht]
\caption{The performance of CityGPT@Beijing on general benchmarks is comparable to that of the baseline.}
\centering
\resizebox{0.38\textwidth}{!}{
\begin{tabular}{lccc} 
\toprule
\textbf{General Benchmark}& \textbf{MMLU}&\textbf{GSM8K} & \textbf{BBH}\\
\midrule
\textbf{Qwen2.5-7B}&  74.15& 80.21& 66.01
\\
\textbf{CityGPT-Qwen2.5-7B}&  74.72$\uparrow$& 77.18$\approx$& 70.03$\uparrow$\\
\midrule
\textbf{LLama3-8B}&  68.33& 79.38& 52.88
\\
\textbf{CityGPT-LLama3-8B}&  56.89& 60.58& 53.92$\uparrow$\\
\midrule
\textbf{ChatGLM3-6B}&  51.97& 57.47& 
34.35\\
\textbf{CityGPT-ChatGLM3-6B}&  52.03$\uparrow$& 56.79$\approx$& 27.51\\

\bottomrule
\end{tabular}}
\label{table:general}
\end{table}

\begin{figure}
    \centering
    \subfigure[Loss on the training data.]{
        \includegraphics[width=0.21\textwidth]{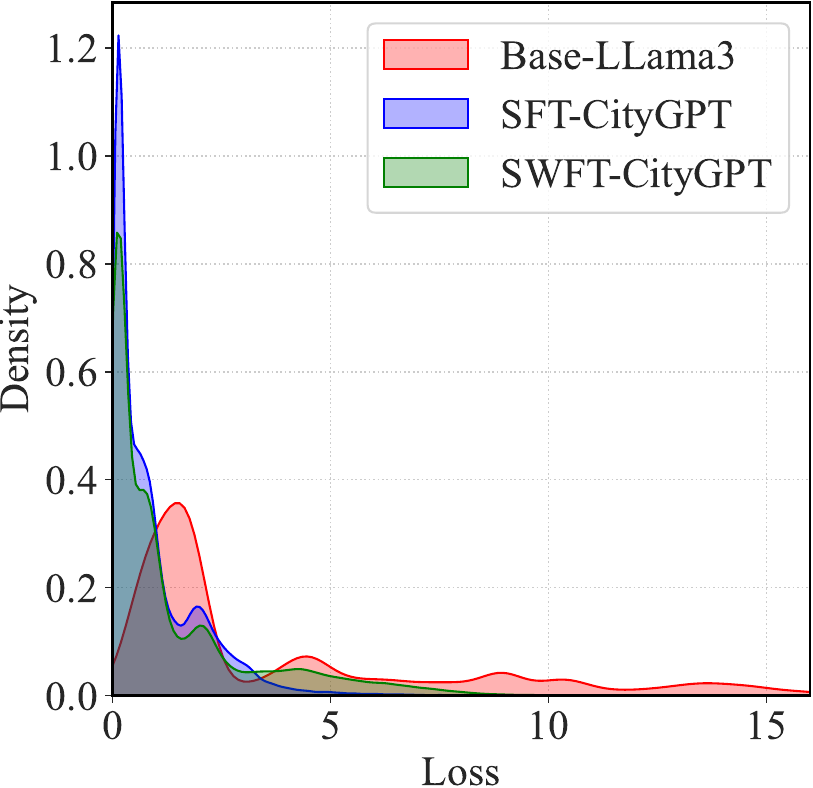}
        \label{figs:loss-comp}
    }
    \subfigure[Performance of models@London.]{
        \includegraphics[width=0.21\textwidth]{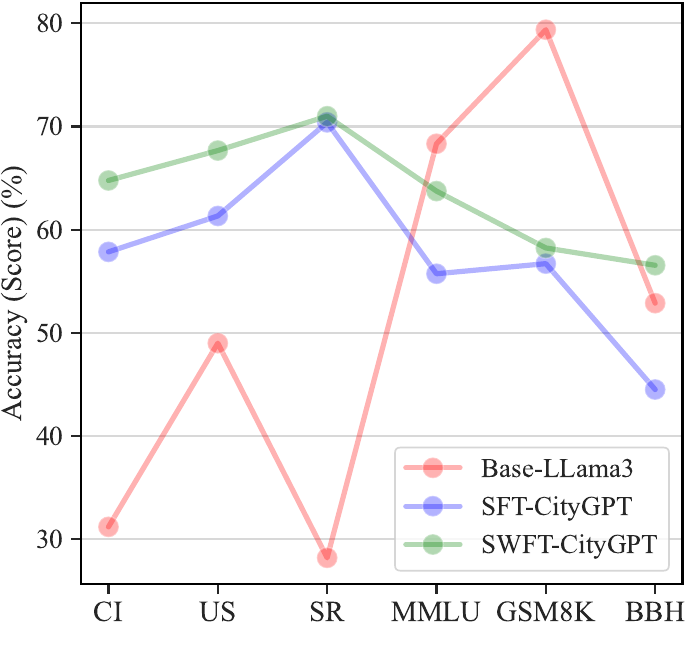}
        \label{figs:loss-improve}
    }
    \caption{Effectiveness of proposed self-weighted tuning.}
    \vspace{-0.3cm}
    \label{fig:loss}
\end{figure}

We conduct exploratory experiments on \textit{SWFT} using LLama3-8B, which experiences the largest decline in general capabilities. As Figure~\ref{figs:loss-comp} shows, the loss range of SFT-CityGPT is significantly narrower than that of the base model, indicating the effectiveness of the fine-tuning process. Notably, the loss for SWFT-CityGPT is larger due to the application of the weighted $f_{exp}$ function in the loss calculation. From Figure~\ref{figs:loss-improve}, we can observe that SWFT-CityGPT significantly improves performance over CityGPT-SFT on both \textit{CityEval} and general benchmark, with gains ranging from 0.8\% to 27\%, demonstrating the effectiveness of the proposed \textit{SWFT} approach. However, it is also evident that CityGPT-SWFT still shows a decrease on the GSM8K task compared to the Base-LLama3 model. This result indicates that while our training strategy helps the model retain its general capabilities on simpler tasks, further improvements are needed for it to sustain comparable performance on more challenging benchmarks.

In summary, after tuning on \textit{CityInstruction} with \textit{SWFT} method, CityGPT demonstrates stronger urban knowledge and task-solving ability while preserving general capabilities.

\subsection{Spatial Transferability and Downstream Task Applicability} \label{exp:app}
In this section, we evaluate the transferability of the spatial knowledge and reasoning ability of CityGPT. As Table~\ref{table:transfer} shows, we divide the results of spatial reasoning into with-context and without-context two groups where each group has 20 questions. Regardless of the base model used, CityGPT@Beijing trained by data from Beijing outperforms the base model significantly in the \textit{CityEval} of other three different cities. The outstanding performance of CityGPT in out-of-domain regions demonstrate that CityGPT indeed learns the general spatial cognition knowledge of urban space which can be transferred between cities. For all models, the performance in the without-context setting is worse than that in the with-context setting, indicating that while \textit{CityInstruction} has enhanced the model's urban space capabilities, there is still room for improvement in handling reasoning tasks.

\begin{table*}
\caption{The performance of CityGPT, trained using data from Beijing, in three another cities: London, NewYork and Paris. `with context' denotes spatial reasoning with context, `w/o context' denotes spatial reasoning without context in \textit{CityEval}.}
\setlength{\tabcolsep}{0.5mm}
\centering
\resizebox{0.98\textwidth}{!}{
\begin{tabular}{lcccccccccccc} 
\toprule
\multirow{2}{*}{\textbf{Model}} & \multicolumn{4}{c}{\begin{tabular}[c]{@{}c@{}}\textbf{Testing}\textbf{@London}\end{tabular}}& \multicolumn{4}{c}{\begin{tabular}[c]{@{}c@{}}\textbf{Testing}\textbf{@NewYork}\end{tabular}}& \multicolumn{4}{c}{\begin{tabular}[c]{@{}c@{}}\textbf{Testing}\textbf{@Paris}\end{tabular}}\\
 & \textbf{CI}$\uparrow$ & \textbf{US}$\uparrow$ & \textbf{with context}$\uparrow$& \textbf{w/o context}$\uparrow$& \textbf{CI}$\uparrow$ & \textbf{US}$\uparrow$ & \textbf{with context}$\uparrow$& \textbf{w/o context}$\uparrow$& \textbf{CI}$\uparrow$ & \textbf{US}$\uparrow$ &\textbf{with context}$\uparrow$& \textbf{w/o context}$\uparrow$\\ 
\midrule
\textbf{Qwen2.5-7B}&  0.303 &  0.480 &  0.270&  0.242&  0.311 &  0.503 &  0.262&  0.250&  0.277 &  0.483 &  0.254&  0.210\\
\begin{tabular}[c]{@{}l@{}}\textbf{Training}\textbf{@Beijing}\end{tabular}&  \textbf{0.511} &  \textbf{0.633} &  \textbf{0.752}& \textbf{0.324}& \textbf{0.483} & \textbf{0.633} & \textbf{0.780}& \textbf{0.354}& \textbf{0.468} & \textbf{0.650} & \textbf{0.764}& \textbf{0.374}\\
\midrule
\textbf{LLama3-8B}& 0.312 & 0.490 & 0.362& 0.202& 0.331 & 0.503 & 0.358& 0.196& 0.265 & 0.463 &0.366& 0.196\\
\begin{tabular}[c]{@{}l@{}}\textbf{Training}\textbf{@Beijing}\end{tabular}& \textbf{0.565} & \textbf{0.653} & \textbf{0.828}& \textbf{0.598}& \textbf{0.543} & \textbf{0.683} & \textbf{0.788}& \textbf{0.570}& \textbf{0.534} & \textbf{0.650} & \textbf{0.836}& \textbf{0.608}\\
\midrule
 \textbf{ChatGLM3-6B}& 0.269 & 0.450 & 0.246 & 0.236 & 0.315 & 0.430 & 0.258 & 0.250 & 0.291 & 0.383 & 0.264 &0.262 \\
 \begin{tabular}[c]{@{}l@{}}\textbf{Training}\textbf{@Beijing}\end{tabular}& \textbf{0.531} & \textbf{0.647} & \textbf{0.814} &\textbf{0.536} & \textbf{0.486} & \textbf{0.663} & \textbf{0.784} & \textbf{0.532} & \textbf{0.509} & \textbf{0.563} & \textbf{0.782} &\textbf{0.548} \\
 \bottomrule
\end{tabular}}
\label{table:transfer}
\end{table*}

The result of downstream \textit{Urban Composite Tasks} group is presented in Table~\ref{table:agent}. CityGPT, acquiring urban spatial knowledge and the skill of spatial reasoning after tuning with \textit{CityInstruction}, outperforms ChatGLM3-6B significantly in all three tasks, including the mobility prediction, trajectory generation and street navigation. It is noted that no task related instruction data is collected and used to train the model during the whole experiment. In the mobility prediction task, the performance of Llama3-70B is competitive with CityGPT. Due to the periodical characteristics of human mobility, Llama3-70B with powerful reasoning capability can observe this regularity and give proper location prediction results by only considering the locations exists in the past trajectory. 
\fix The results demonstrate that when the model is trained with knowledge-based QA and navigation tasks, it learns the local knowledge of the urban space from the QA and the mobility patterns of the population from navigation instructions. Both of these are valuable for various downstream tasks. While this may not directly enhance the model's personalized capabilities for tasks like mobility prediction and trajectory generation, a deeper understanding of urban spaces and context can still improve its performance on these tasks. \color{black}
\begin{table*}
    \caption{CityGPT-ChatGLM3-6B performs well on \textit{Composite Task} without any further task-related fine-tuning. In the table, \textbf{bold} denotes the best results, \underline{underline} denotes the second best results.}
    \centering
    \resizebox{0.9\textwidth}{!}{
    \begin{tabular}{lcccccc} 
    \toprule
    \textbf{Tasks@Beijing-Wudaokou} & \multicolumn{2}{c}{\textbf{Mobility Prediction}} & \multicolumn{2}{c}{\textbf{Trajectory Generation}} & \multicolumn{2}{c}{\textbf{Spatial Navigation}} \\
     & \multicolumn{1}{l}{\textbf{Acc(multi) $\uparrow$}} & \multicolumn{1}{l}{\textbf{Acc(gen) $\uparrow$}} & \multicolumn{1}{l}{\textbf{Radius(JSD) $\downarrow$}} & \multicolumn{1}{l}{\textbf{Distance(JSD)} $\downarrow$} & \multicolumn{1}{l}{\textbf{Steps} $\downarrow$} & \multicolumn{1}{l}{\textbf{Success Rate} $\uparrow$} \\ 
    \midrule
    \textbf{ChatGLM3-6B} & 0.25 & 0.12 & 0.473 & 0.416 & 20.83 & 0.35 \\
    \textbf{CityGPT-ChatGLM3-6B} & \textbf{0.52} & \underline{0.46} & \textbf{0.451} & \textbf{0.380} & \textbf{15.32}~ & \textbf{0.55} \\
    \textbf{Llama3-70B} & \underline{0.45} & \textbf{0.50} & \underline{0.455} & \underline{0.389} & \underline{18.95} & \underline{0.44} \\
    \bottomrule
    \end{tabular}
    }
    \label{table:agent}
\end{table*}

\subsection{Instruction Training Data Ablation Study} \label{exp:simpleshot}
\begin{figure}
\centering
\includegraphics[width=0.46\textwidth]{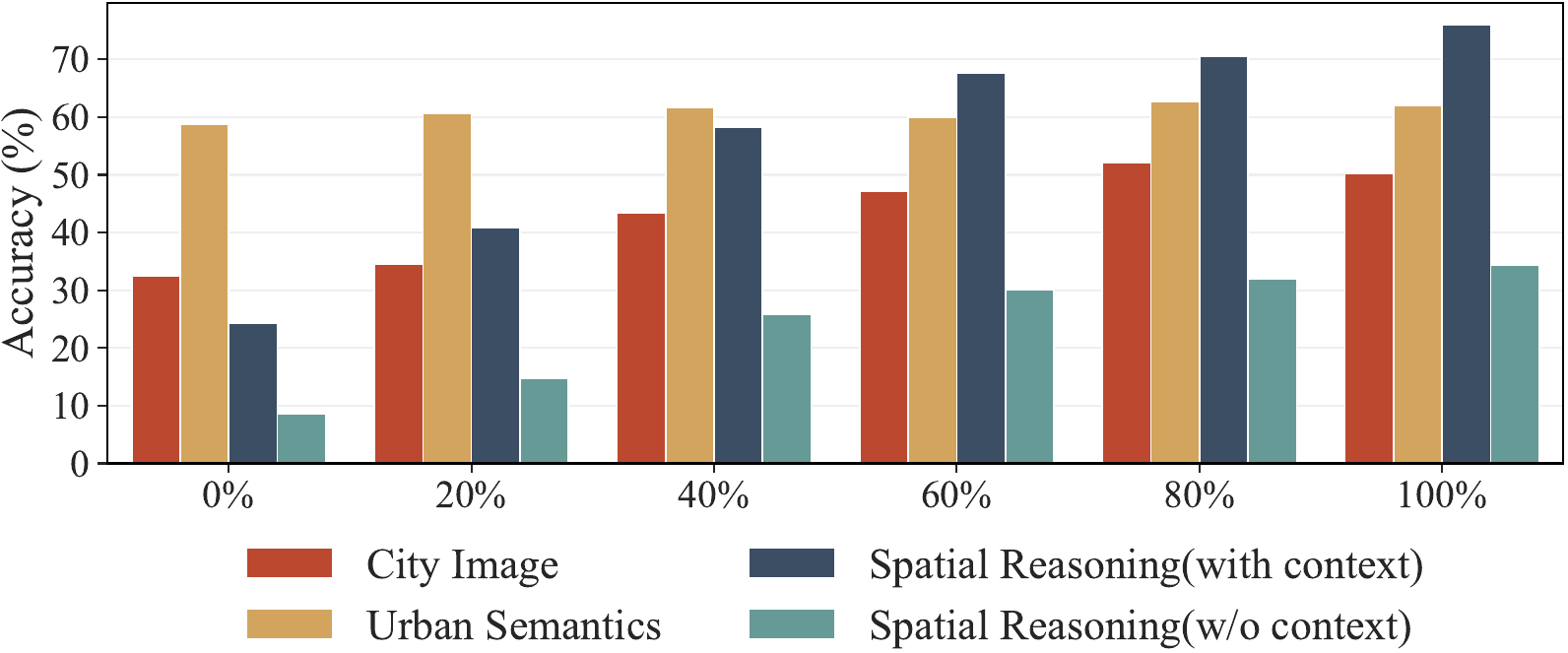}
\caption{\fix The performance of CityGPT improves as the volume of the instruction dataset increases.}
\vspace{-0.3cm}
\label{fig:ablation_study:data_volume}
\end{figure}

\begin{figure}
    \centering
    \includegraphics[width=0.46\textwidth]{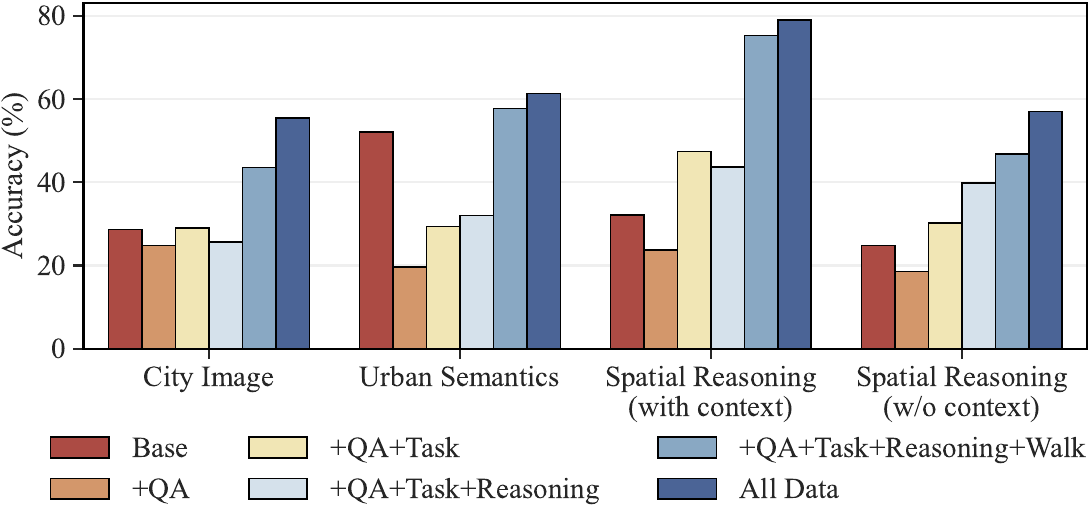}
    \caption{The performance of CityGPT increases when different sub-datasets are added, where `base’ denotes the base model, `+’ denotes the model is trained on added dataset, `Task’ denotes task instruction data.}
    \label{fig:ablation_study}
\end{figure}

In this section, we study the influence of the data composition in \textit{CityInstruction} in Figure~\ref{fig:ablation_study:data_volume}, Figure~\ref{fig:ablation_study}, 
and Table~\ref{tab:data:simple}. To study the effects of data size, we split the whole \textit{CityInstruction} into five equal parts and fine-tune Qwen2.5-7B with data from only 1 part to all the 5 parts. As Figure~\ref{fig:ablation_study:data_volume} shows, the performance of all the tasks increase with more data are utilized. We find that \textit{Urban Semantics} task group requires only a modest amount of data to achieve a relatively high level of performance, whereas the more challenging \textit{Spatial Reasoning} task necessitates significantly larger quantities of data to attain a comparable high level of performance. 

\begin{table}[ht]
\centering
\caption{\fix Comparison of our data with SimpleData.}
\begin{tabular}{lccc} 
\toprule
\textbf{Model@Beijing}& \textbf{CI} & \textbf{US} & \textbf{SR} \\ 
\midrule
LLama3-8B & 0.286 & 0.520 & 0.285 
\\
LLama3-8B-SimpleData & 0.219 & 0.480 & 0.191 
\\
LLama3-8B-SimpleData2 & 0.259 & 0.467 & 0.260 \\
\textbf{CityGPT-8B-OurData} & \textbf{0.554}& \textbf{0.680}& \textbf{0.708}\\
\bottomrule
\end{tabular}
\label{tab:data:simple}
\end{table}

\fix Besides, we compare our \textit{CityInstruction} with the simple tuning data to demonstrate the effectiveness of data construction in our framework. As shown in Table~\ref{tab:data:simple}, without carefully designed instruction-tuning data, simple fine-tuning on naive static urban knowledge does not enable LLMs to effectively understand the urban space. In the table, all the models utilize standard SFT, with the only difference being the training data. "SimpleData" refers to the direct conversion of geospatial data into conversational format for training the LLM, while "SimpleData2" indicates the mixing of SimpleData with the same general text instruction data used in CityGPT. Our findings show that CityGPT, when trained with carefully designed instruction data, outperforms LLMs trained solely on simple geospatial data. Although learning from the simple data provides some performance improvements in certain cases, it can also negatively impact the LLM in others. 

Furthermore, we discuss the effects of different types of data in \textit{CityInstruction} and present the results in Figure~\ref{fig:ablation_study}. All Data' yields the best performance, showing that single datasets are insufficient and mixing multiple datasets is key to superior results.

\fix Finally, we present the in-context learning (5-shot) results for three representative baselines—LLama3.1-405B and LLama3-8B—in Beijing in the Table~\ref{tab:incontext}. As shown in the table, compared to the zero-shot setting, the performance of the baselines in the 5-shot setting exhibits varying degrees of improvement, ranging from 8\% to 15\%. This highlights the importance of providing examples to help LLMs become familiar with the task paradigm and instructions. However, despite the performance gains in the 5-shot setting, these baselines still lag behind CityGPT in the zero-shot setting in the vast majority of cases. We also provide the results of Qwen in appendix.\color{black}
\begin{table}
\centering
\caption{\fix Performance of \textit{CityGPT} with zero-shot settings and general LLMs with few-shot settings in \textit{CityEval}.}
\label{tab:incontext}
\begin{tabular}{lccc} 
\toprule
\textbf{Model@Beijing}  & \textbf{CI} & \textbf{US} & \textbf{SR} \\ 
\midrule
LLama3.1-405B-zeroshot& 0.417 & 0.587 & 0.414 
\\
LLama3.1-405B-5shot& 0.451 & 0.663& 0.468 
\\ 
\midrule
LLama3-8B-zeroshot & 0.286 & 0.520 & 0.285 
\\
LLama3-8B-5shot & 0.330 & 0.566 & 0.262 
\\
\textbf{CityGPT-8B}& \textbf{0.554}& \textbf{0.680}& \textbf{0.708}\\
\bottomrule
\end{tabular}
\end{table}

\section{Related Work}

\textbf{Large Language Models.} 
Since the publication of ChatGPT~\cite{ChatGPT}, LLMs~\cite{zhao2023survey} presents impressive language generation and reasoning capabilities in many challenging tasks~\cite{achiam2023gpt,feng2025survey}. While LLMs~\cite{touvron2023llama, zeng2022glm} are trained on the massive online web text data, they usually underperform in many specialized fields of real-life. Thus, researchers design various mechanisms to enhance the capability of general LLMs in specific domains, like BloombergGPT~\cite{wu2023bloomberggpt} , Med-Plam~\cite{singhal2023large}, and Llemma~\cite{azerbayev2023llemma}. In the geospatial field, K2~\cite{deng2024k2} is designed to answer geographical knowledge questions after learning scientific literature. Similar to K2, Wang et al.~\cite{wang2023optimizing} finetune ChatGLM to answer questions about urban renewal. Different from these domain specific QA models for text based literature, our method is designed for understanding physical urban space whose original format is not text and conducting spatial reasoning to solve real urban tasks.

\noindent
\textbf{Language and Urban Space.} 
While urban space is usually described by the accurate digital map, researchers have explored various methods~\cite{liu2023urbankg,ding2024understanding,huang2022ernie,li2023geolm,balsebre2024lamp,li2022spabert,feng2025agentmove,feng2025survey} to grounding the natural language with the geospatial entity. Huang et al.~\cite{huang2022ernie} propose utilize online user searching logs to construct heterogeneous graph and sample random walks to construct the sequence data with geospatial entity. Li et al.~\cite{li2022spabert,li2023geolm} use geospatial entity from the Wikipedia to align with the same entity in map. However, the former relies on the large scale private user logs and the latter can only cover limited entities mentioned in the Wikipedia. Recently, Balsebre et al.~\cite{balsebre2024lamp} propose LAMP to injecting PoIs into LLMs with RAG for better PoI recommendation, which is similar to part of the questions about PoIs in our \textit{CityQA}. Unlike LAMP, our method considers a broader range of questions and entities by drawing inspiration from GeoQA~\cite{punjani2018template,mai2021geoqa} and we introduce routing based instruction data and spatial reasoning instructions as new ways to grounding the urban spatial knowledge with language. 

\noindent
\textbf{Spatial Cognition and Reasoning.}  
With explicit function zones in the brain like place cells and border cells, human build a cognitive map~\cite{epstein2017cognitive, farzanfar2023cognitive} in the mind to recognize the physical world and navigate in it. Following this concept, researchers try to evaluate the spatial reasoning and navigation capability of LLMs in abstract environments~\cite{yamada2023evaluating,momennejad2024evaluating,zha2025enable}. However, their experiment are conducted in the idealized and non-realistic environments like grids which is far from the real-life urban space. Besides, many studies~\cite{roberts2023gpt4geo,gurnee2023language} propose to evaluate the spatial knowledge of LLMs. However, most of them pay attention to the coordinates based problems and only conduct national scale evaluations. In this paper, we follow the concept of the image of the city~\cite{lynch1964image} and spatial cognitive map to construct the \textit{CityEval} benchmark. As the first systematic evaluation benchmark for urban spatial capability of LLMs, it covers diverse aspects of urban space and sheds light for effectively evaluating the intelligence and utility of LLMs for urban system.

\section{Conclusion}
In this paper, we propose a systematic framework for evaluating and enhancing the capability of LLMs on understanding urban space and solving related urban tasks. For effectively evaluating the capability of LLMs, we construct \textit{CityEval} which comprehensively considering various aspects of urban space. To enhance the capability of smaller LLMs, we construct a diverse instruction tuning dataset \textit{CityInstruction} with human-like spatial experience data and enhanced spatial reasoning problem data via simulation. Finally, we propose a \textit{self-weighted fine-tuning} method that incorporates data quality into the loss function to mitigate the forgetting problem while enhancing the domain-specific capabilities of LLMs. This approach enables robust and stable learning using ordinary data without the need for costly, meticulous filtering, ensuring resilience against the influence of noisy data. In the future, we plan to validate the framework in more cities around the world to further demonstrate its effectiveness. We will also explore adding data from other modalities, such as remote sensing and street view images, to further enhance the capability of foundation models.

\bibliographystyle{ACM-Reference-Format}
\balance
\bibliography{references}

\appendix
\appendix
\section{Appendix}
\subsection{Comparison of CityEval with other Benchmarks}
 Table~\ref{tab:eval:comp} presents a comparison of our CityEval with several related benchmarks on geospatial and urban knowledge. As shown in the table, CityEval covers a wider range of task types and provides detailed evaluations for three cities. Most existing benchmarks focus solely on assessing the spatial knowledge of LLMs, which is only one aspect of CityEval. In contrast, CityEval includes high-level reasoning questions and multiple real-world downstream application tasks.  Detailed content of four modules in \textit{CityEval} are introduced as follows.
Detailed statistic information for \textit{CityEval} in Beijing can refer to Table~\ref{tab:eval-statis}.

\begin{table*}
\centering
\caption{\fix Comparison of our CityEval with several related benchmarks on geospatial and urban knowledge.}

\begin{tabular}{ccccccc} 
\toprule
 & Types & Content & Instances & Source & Scale & Format \\ 
\midrule
GeoBench~\cite{deng2024k2} & 2 task & GeoScience Exam & \textasciitilde{}2K & NPEE, APTest & - & multi-choice \\
GPT4GEO~\cite{roberts2023gpt4geo} & 19 tasks & limited task for city & \textasciitilde{}100 & GeoNames, Google Map & World & generate \\
Space\&Time~\cite{gurnee2023language} & 1 task & POI coodinates & \textasciitilde{}20K & NYC OpenData POI & NYC & regression \\
CityEval-Ours & 39 tasks & \begin{tabular}[c]{@{}c@{}}CityImage, UrbanSemantics, \\Spatial Reasoning\end{tabular} & \textasciitilde{}15K & Open Street Map & \begin{tabular}[c]{@{}c@{}}Beijing, Paris, \\NewYork, London\end{tabular} & multi-choice \\
\bottomrule
\end{tabular}
\label{tab:eval:comp}
\end{table*}

\begin{table}
    \caption{Statistics of different task groups of \textit{CityEval} in Beijing, where Composite Tasks are conducted in Beijing-Wudaokou, and Acc. denotes accuracy, SR. denotes success rate.}
    \centering
    \begin{tabular}{lccc} 
    \toprule
    \textbf{Task Group} & \textbf{\#Insts.} & \textbf{\#Tasks} & \textbf{Metrics} \\ 
    \hline
    \textbf{City Image} & 650& 13 & Acc \\
    \textbf{Urban Semantics} & 300& 6 & Acc \\
    \textbf{Spatial Reasoning} & 1000& 20 & Acc \\
    \textbf{Composite Tasks} & 321 & 3 & SR, etc. \\
    \bottomrule
    \end{tabular}
    \label{tab:eval-statis}
\end{table}

\subsection{\textit{SWFT} Validation in Other Cities}
\begin{table}[ht]
\caption{The performance of Models using different tuning method on \textit{CityEval} and general benchmarks.}
\centering
\resizebox{0.48\textwidth}{!}{
\begin{tabular}{lllllccc} 
\toprule
 \textbf{City}&\textbf{Model}&    \textbf{CI}&\textbf{US}&\textbf{SR}&\textbf{MMLU}&\textbf{GSM8K} & \textbf{BBH}\\
\midrule
\textbf{Beijing}&\textbf{Base-LLama3}&     0.286 &0.520 &0.285 &68.33& 79.38& 52.88
\\
 &\textbf{SFT-CityGPT}&     0.554 &0.680 &0.708 &56.89& 60.58& 53.92
\\
 & \textbf{SWFT-CityGPT}& 0.605$\uparrow$ & 0.670$\approx$ & 0.709$\approx$ & 60.51$\uparrow$& 57.92$\approx$&55.76$\uparrow$
\\
\midrule
 \textbf{NewYork}&\textbf{Base-LLama3}&     0.331 &0.503 &0.277 &68.33& 79.38& 52.88\\
 &\textbf{SFT-CityGPT}&     0.574 &0.650 &0.712 
&55.12& 59.44& 53.37
\\
 & \textbf{SWFT-CityGPT}& 0.654$\uparrow$& 0.673$\uparrow$& 0.704$\approx$& 63.72$\uparrow$& 61.56$\uparrow$&56.29$\uparrow$\\
  \bottomrule
\end{tabular}}
\label{table:swft_app}
\end{table}

This section presents comparative results for models trained using different methods in Beijing and NewYork. From the Table~\ref{table:swft_app}, we observe that models trained using SWFT outperform those obtained through standard SFT training.

\subsection{Effects of the LLM Size}
\begin{figure}
\centering
\includegraphics[width=0.45\textwidth]{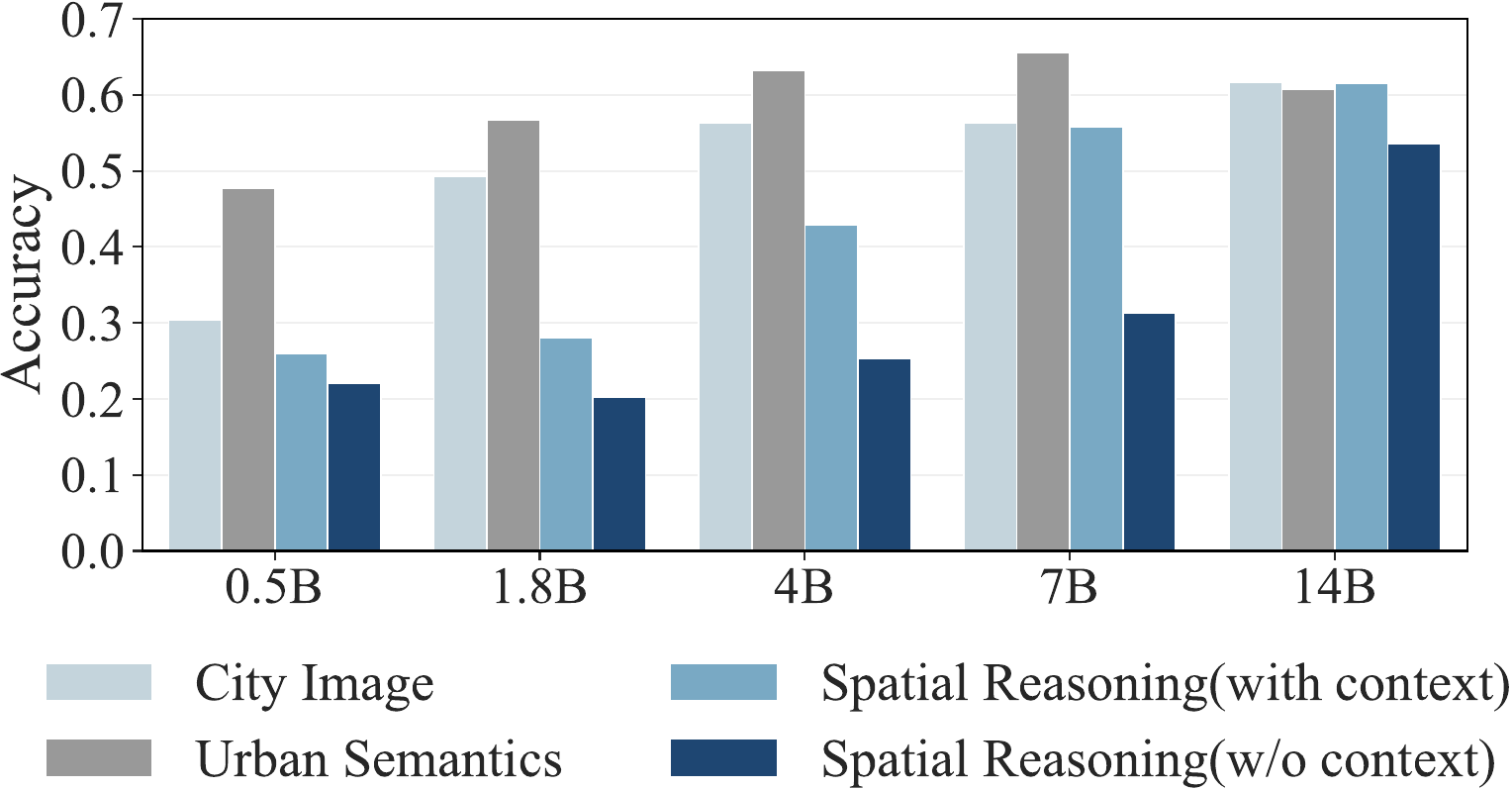}
\caption{ The performance of CityGPT improves as the parameters of Qwen1.5 increase, from 0.5B to 14B.}
\label{fig:ablation_study:model_size}
\end{figure}

In this section, we study the effects of model in Figure~\ref{fig:ablation_study:model_size}. We select the Qwen1.5 with 0.5B-14B models to investigate the influence of the model size. Due to limited computing resources in our experiments, all Qwen1.5 models were trained with full fine-tuning settings, except for Qwen1.5-14B, which was trained using LoRA settings. As Figure~\ref{fig:ablation_study:model_size} shows, we can observe that the challenging \textit{Spatial Reasoning} task require LLMs with larger parameters while the \textit{Urban Semantics} task group can be handled well with smaller LLMs. Furthermore, CityGPT-Qwen1.5-14B outperforms CityGPT-ChatGLM3-6B with more than 20\% in the \textit{Spatial Reasoning} task. Although Qwen1.5-14B shows only minimal performance gains and even declines in certain tasks compared to smaller models, we believe this may be due to the LoRA training mechanism.

\subsection{Additional Results of SimpleData}
Extending our exploration from Section~\ref{exp:simpleshot}, this section includes the SimpleData and 5-shot results for Qwen2.5-7B in Table~\ref{tab:data:simple_app} and Table~\ref{tab:incontext_app}, which are identical to those for LLama3-8B. Our CityGPT significantly outperforms models trained with simple urban geographic data and baseline with 5-shot setting.
\begin{table}[ht]
\centering
\caption{\fix Comparison of our data with SimpleData on Qwen2.5-7B.}
\vspace{-0.3cm}
\begin{tabular}{lccc} 
\toprule
\textbf{Model@Beijing}& \textbf{CI} & \textbf{US} & \textbf{SR} \\ 
\midrule
Qwen2.5-7B& 0.325 & 0.587 & 0.164 
\\
Qwen2.5-7B-SimpleData& 0.320 & 0.583 & 0.252 
\\
Qwen2.5-7B-SimpleData2& 0.312 & 0.573 & 0.191 \\
\textbf{CityGPT-7B-OurData}& \textbf{0.502}& \textbf{0.620}& \textbf{0.552}\\ 
\bottomrule
\end{tabular}
\vspace{-0.3cm}
\label{tab:data:simple_app}
\end{table}

\begin{table}
\centering
\caption{\fix Performance of \textit{CityGPT} with zero-shot settings and Qwen2.5-7B with few-shot settings in \textit{CityEval}.}
\vspace{-0.3cm}
\label{tab:incontext_app}
\begin{tabular}{lccc} 
\toprule
\textbf{Model@Beijing}  & \textbf{CI} & \textbf{US} & \textbf{SR} \\ 
\midrule
Qwen2.5-7B-zeroshot& 0.325 & 0.587 & 0.164 
\\
Qwen2.5-7B-5shot& 0.368 & 0.596 & 0.279 
\\
\textbf{CityGPT-7B}& \textbf{0.502}& \textbf{0.620} & \textbf{0.552}\\ 
\bottomrule
\vspace{-0.3cm}
\end{tabular}
\end{table}

\subsection{Extended Results of Spatial Transferability}
Following up on the results in Section~\ref{exp:app}, we also conduct urban transferability assessments for CityGPT-7B, trained with data from different cities. The results can be viewed in Table~\ref{table:transfer_app}. Models trained with data from CityA have achieve excellent results in evaluation of CityB.
\begin{table}
\caption{The performance of CityGPT-7B, trained using data from CityA, evaluated using data from CityB. }
\vspace{-0.3cm}
\setlength{\tabcolsep}{0.5mm}
\centering
\resizebox{0.48\textwidth}{!}{
\begin{tabular}{llccc} 
\toprule
 \textbf{Task}&& \begin{tabular}[c]{@{}c@{}}\textbf{Training}\textbf{@London}\end{tabular}& \begin{tabular}[c]{@{}c@{}}\textbf{Training}\textbf{@NewYork}\end{tabular}& \begin{tabular}[c]{@{}c@{}}\textbf{Training}\textbf{@Paris}\end{tabular}\\ 
\midrule
 \textbf{Beijing}&\textbf{CI}&  0.482 &  0.460 &  0.437 
\\
 & \textbf{US}& 0.543 & 0.563 & 0.530 
\\
 &\textbf{SR}&  0.584 & 0.598 & 0.588 
\\
\midrule
 \textbf{London}&\textbf{CI}& /& 0.506 & 0.434 
\\
 & \textbf{US}& /& 0.567 & 0.567 
\\
 &\textbf{SR}& /& 0.596 & 0.605 
\\
\midrule
  \textbf{NewYork}&\textbf{CI}& 0.465 & /& 0.486 
\\
 & \textbf{US}& 0.573 & /& 0.590 
\\
  &\textbf{SR}& 0.576 & /& 0.626 
\\
 \bottomrule
\end{tabular}}
\vspace{-0.3cm}
\label{table:transfer_app}
\end{table}

\begin{figure*}[!hbtp]
    \centering
    \includegraphics[width=1.0\textwidth]{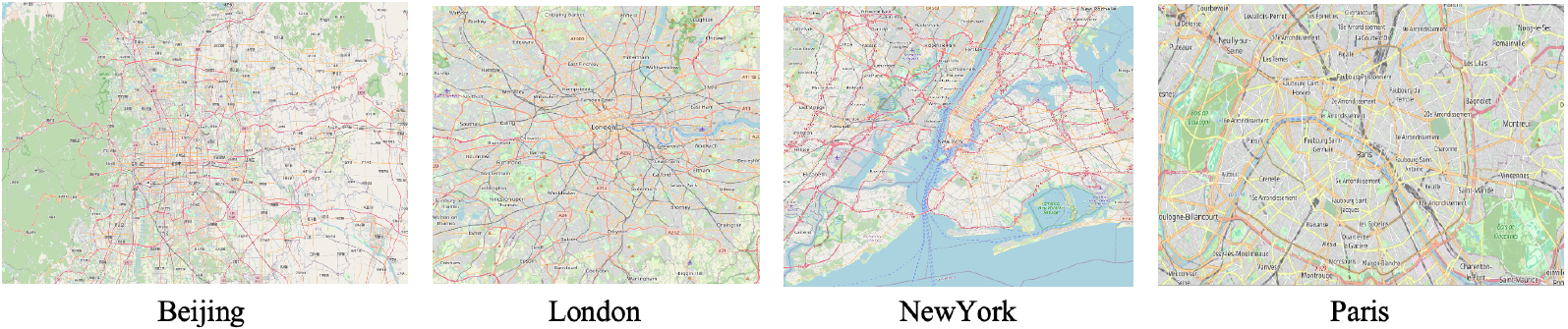}
    \caption{Spatial range of evaluated cities in the experiments.}
    \label{fig:map_all}
\end{figure*}

\subsection{Data Statistics}
 We present the spatial range of the four evaluated regions in Figure~\ref{fig:map_all} and the constructed \textit{CityInstruction} data for each city in Table~\ref{table:citydata-info}. 

\begin{table}[h]
\centering
\caption{ Basic statistical information of \textit{CityInstruction}. \textit{CityReasoning} data for the four cities are the same which are generated from SanFrancisco.}
\label{table:citydata-info}
\resizebox{0.47\textwidth}{!}{
\begin{tabular}{llccc} 
\toprule
\textbf{City} & \textbf{Dataset} & \textbf{Instance} & \textbf{Length/Token} & \textbf{Rounds} \\ 
\hline
\multirow{2}{*}{\begin{tabular}[c]{@{}l@{}}\textbf{General}\\\textbf{Instruction Data}\end{tabular}} & \textbf{Spatial} & 3488 & 283 & 1 \\
 & \textbf{Chat} & 41866& 745& 2.32\\ 
\hline
\multirow{3}{*}{\begin{tabular}[c]{@{}l@{}}\textbf{CityInstruction@}\textbf{Beijing}\end{tabular}}& \textbf{CityWalk} & 30000& 593& 1\\
 & \textbf{CityQA} & 48551& 105& 1\\
 & \textbf{CityReasoning} & 7992& 688& 1.13\\ 
\hline
\multirow{3}{*}{\begin{tabular}[c]{@{}l@{}}\textbf{CityInstruction@}\textbf{London}\end{tabular}}& \textbf{CityWalk} & 30000& 537& 1\\
 & \textbf{CityQA} & 48691& 88& 1\\
 & \textbf{CityReasoning} & 7992& 688& 1.13\\ 
\hline
\multirow{3}{*}{\begin{tabular}[c]{@{}l@{}}\textbf{CityInstruction@}\textbf{NewYork}\end{tabular}}& \textbf{CityWalk} & 30000& 607& 1\\
 & \textbf{CityQA} & 48484& 95& 1\\
 & \textbf{CityReasoning} & 7992& 688& 1.13\\ 
\hline
\multirow{3}{*}{\begin{tabular}[c]{@{}l@{}}\textbf{CityInstruction@}\textbf{Paris}\end{tabular}}& \textbf{CityWalk} & 29128& 645& 1\\
 & \textbf{CityQA} & 41546& 108& 1\\
 & \textbf{CityReasoning} & 7992& 688& 1.13\\
\bottomrule
\end{tabular}}
\end{table}

\subsection{Description of Urban Composite Tasks}
\textbf{Mobility Prediction:} 
The model is required to predict the next PoI of a person based on his/her previous trajectory. Information provided include: 1) previous trajectory of the person (a series of trajectory items formatted as [poi name, visiting time]), 2) the visiting time of the PoI that requires prediction, 3) 9 candidate prediction PoIs. Note that the trajectory data used in this experiment are all from real world users. Ground truth of each question is extracted from the trajectory dataset.

\noindent
\textbf{Trajectory Generation:}
The model generates a trajectory based on a virtual agenda generated by GPT-3.5-turbo-1106. First, 250 templates of virtual agenda, formatted as a series of [time, action] items, is generated with GPT-3.5. Then, the model is required to assign a possible poi to each [time, action] item, which creates a trajectory with each point in time assigned with a corresponding PoI.
The performance of the model is evaluated by comparing the virtually generated trajectories with the real-word ones. Specifically, we use JSD to measure the similarity between the mobility pattern distributions of generated trajectory and real trajectory data.

\noindent
\textbf{Spatial Navigation:}
The model is required to make step by step navigation from one AoI to another. At each step, the model is provided with: 1) hint about its current position(denoted by the two PoIs closest to its current position), 2) name of the destination AoI B, 3) candidate choices of navigation lanes (extracted from the map simulating platform and formatted as [road name, direction]).
The choice of the model among the candidates will lead to a position update towards the next crossroad along the chosen road and its corresponding direction. The task is deemed successful if the model is able to navigate itself to a position that is within a threshold distance( 500m in our experiment) to the destination AoI B in 30 steps.
For evaluation, the models are tested on 21 navigation tasks designed to be finished within a minimum of 1,3,6 steps and an average of 4.5 steps. 
Metrics are introduced as below.

\begin{itemize}[leftmargin=1.5em,itemsep=0pt,parsep=0.2em,topsep=0.0em,partopsep=0.0em]
\item  \textbf{Acc(multi)}: prediction accuracy when N candidate PoIs are provided.
\item  \textbf{Acc(gen)}: prediction accuracy when no candidate PoIs are given.
\item  \textbf{Radius(JSD)}: radius of gyration, which represents the spatial range of the user’s daily activities. 
\item  \textbf{DailyLoc(JSD)}: daily visited locations, which is calculated as the number of visited locations per day for user.
\item  \textbf{Steps}: average step that the model took to finish the navigation task. Note that the step is counted as 30 (maximum rounds of navigation) for cases where the model failed to navigate to the destination.
\item  \textbf{Success Rate}: probability that the model successfully navigates to the specified location.
\end{itemize}

\subsection{Hyper-parameter Settings} \label{sec:para}
\begin{table}[h]
\centering
\caption{Hyper-parameters for fine-tuning.}
    \label{table:para}
    \begin{tabular}{cccc} 
    \toprule
    \textbf{Hyper-parameter} & \textbf{Value} & \textbf{Hyper-parameter} & \textbf{Value} \\
    Learning Rate & 1e-5 & Max Sequence Length & 4096 \\
    Batch Size Per Device & 1 & Epoch~ & 1 \\
    Gradient Acc Steps & 16 & LR Scheduler & cosine \\
    \bottomrule
    \end{tabular}
\end{table}
To reduce the training cost, we choose to fine-tune the chat version of general LLMs. We utilize accelerate~\cite{accelerate} with the full training mode to fine-tune the LLMs in a server with 4 A800 GPUs. We try to train ChatGLM3-6B~\cite{zeng2022glm}, Qwen series~\cite{qwen}, and LLama3-8B~\cite{touvron2023llama} in our experiments. 
During the experiments, we find that the influence of parameter settings is far less than the influence of data quality. Thus, we fix the training parameters in most of the experiments and leave the work of best parameter search in the future. The core training parameters settings are shown in Table~\ref{table:para}.

\end{document}